\documentclass[10pt,twocolumn,letterpaper]{article}

\usepackage[pagenumbers]{cvpr} 

\newcommand{\new}[1]{\textcolor{black}{#1}}

\definecolor{cvprblue}{rgb}{0.21,0.49,0.74}
\usepackage[pagebackref,breaklinks,colorlinks,allcolors=cvprblue]{hyperref}
\usepackage[accsupp]{axessibility}

\makeatletter
\renewcommand{\paragraph}[1]{\noindent\textbf{#1}\ }
\makeatother

\newcommand{\methodname}{POLAR\xspace}

\def\paperID{7636} 
\def\confName{CVPR}
\def\confYear{2025}

\title{Improving Personalized Search with Regularized Low-Rank Parameter Updates}
\author{Fiona Ryan$^{1,2*}$, Josef Sivic$^{2,3}$, Fabian Caba Heilbron$^2$, Judy Hoffman$^1$, James M. Rehg$^4$, Bryan Russell$^2$\\
$^1$Georgia Tech, $^2$Adobe Research, $^3$CIIRC CTU, $^4$UIUC \\
}

\begin{document}
\maketitle

{\let\thefootnote\relax\footnote{{*Work partially done during internship at Adobe Research.}}}

\begin{abstract}
Personalized vision-language retrieval seeks to recognize new concepts (\eg, ``my dog Fido'') from only a few examples. This task is challenging because it requires not only learning a new concept from a few images, but also integrating the personal and general knowledge together to recognize the concept in different contexts. In this paper, we show how to effectively adapt the internal representation of a vision-language dual encoder model for personalized vision-language retrieval. We find that regularized low-rank adaption of a small set of parameters in the language encoder's final layer serves as a highly effective alternative to textual inversion for recognizing the personal concept while preserving general knowledge. Additionally, we explore strategies for combining parameters of multiple learned personal concepts, finding that parameter addition is effective. To evaluate how well general knowledge is preserved in a finetuned representation, we introduce a metric that measures image retrieval accuracy based on captions generated by a vision language model (VLM). Our approach achieves state-of-the-art accuracy on two benchmarks for personalized image retrieval with natural language queries -- DeepFashion2 and ConCon-Chi -- outperforming the prior art by $4\%-22\%$ on personal retrievals.
\end{abstract}    
\vspace{-2em}
\section{Introduction}
\label{sec:intro}

Personalizing a vision-language retrieval model (PerVL) aims to adapt a pretrained vision-language dual encoder model (\eg, CLIP~\cite{radford2021clip}) to recognize new concepts (\eg, ``my dog Fido'') from just a few examples~\cite{cohen2022my}. This task is important for search applications that need to identify concepts missing from the pretrained model's knowledge, such as searching one's personal photo library for a specific person, object, or pet. PerVL is challenging because it requires not only learning a new concept from a few visual examples, but also reasoning about the personal concept and general knowledge together to retrieve the concept in different contexts. For instance, searching for ``my dog Fido catching a frisbee" requires both personal knowledge (``Fido") and general knowledge (``catching a frisbee").

\begin{figure}[t]
  \centering
\includegraphics[width=1.0\linewidth]{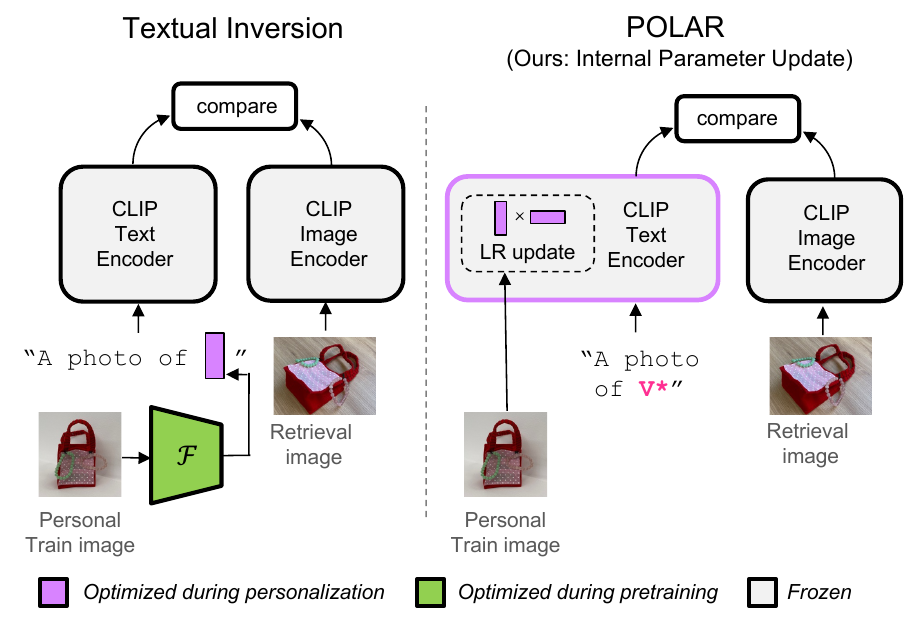}

   \caption{{\em Left:} Prior works use a pretrained textual inversion network ($\mathcal{F}$) to compute a pseudo-token to represent a new concept, which may be further optimized during personalization. {\em Right:} We present POLAR, which represents new concepts as a small low-rank parameter update within the text encoder. Instead of inserting a learned pseudo-token to the text query, we use a fixed vocabulary token \textit{V*}.  We show that our method is effective at recognizing the personal concept from a few examples while retaining the model's general knowledge, and does not require large scale pretraining.}
   \vspace{-1em}
   \label{fig:teaser}
\end{figure}

Recent approaches for PerVL use \textit{textual inversion} to learn a pseudo-text token (\eg, ``{\em V*}'') for each new personal concept by inverting from the target training images to the token~\cite{cohen2022my, baldrati2023zero} (illustrated in Figure~\ref{fig:teaser} (left)). These pseudo-text tokens require updating only a minimal set of parameters, and they can be used within a natural language query as input to the language encoder (\eg, ``{\em V* catching a frisbee}''). 
Typically, these approaches train a textual inversion network on large-scale data to predict a token representing a new concept, which may then be further optimized during personalization. 
While these approaches avoid updating the model's internal parameters, preventing the ``overriding'' of its general knowledge, their ability to represent the personal concept is limited to the single input text token. 

Furthermore, the token affects the entire text embedding process and can interfere with the language encoder's general knowledge. Consequently, textual inversion approaches often struggle to combine personalized and general knowledge across different datasets and can be time-consuming to optimize online due to the need for backpropagation through the full language encoder~\cite{cohen2022my, yeh2023meta}, or require large-scale pretraining~\cite{cohen2022my, baldrati2023zero, yeh2023meta}. 

To address these shortcomings, we focus on updating the internal representation of CLIP's language encoder~\cite{radford2021clip} using only a few training examples (illustrated in Figure~\ref{fig:teaser} (right)). Updating the internal representation is challenging because the model must balance learning the personalized concept from limited examples while preventing catastrophic forgetting of its prior knowledge. Additionally, the training data does not inherently encourage the model to retain general knowledge. Our goal is to make this update efficiently, without relying on additional training data. 

Our contributions are fourfold. First, inspired by recent advances in text-to-image generation~\cite{ruiz2023dreambooth,kumari2023multi, ruiz2024hyperdreambooth, ham2024personalized}, we show that CLIP's language encoder can be updated to learn a personalized concept from a few examples while retaining its general knowledge. We find that regularized low-rank adaptation (LoRA) finetuning~\cite{hu2021lora} of the language encoder's last layer effectively balances the trade-off between learning the personalized concept and avoiding catastrophic forgetting. Our resulting method \textbf{\methodname} (PersOnalized Low-rank Adaptation for Retrieval) learns a low-rank parameter set that is minimal and comparable in size to the pseudo-text tokens used in textual inversion. Furthermore, we leverage LoRA's specific structure to introduce a regularization strategy that eliminates the need for additional training examples or regularization prompts. Unlike previous methods, our approach does not rely on any components pre-trained on large-scale data, allowing for seamless parameter updates using only a few training examples. 

Second, we introduce a new evaluation metric to assess how well general knowledge is preserved in our finetuned representation, using captions generated by a vision-language model (VLM). We find that our approach effectively maintains general knowledge. 
Third, we explore different strategies for combining learned representations for different personal concepts to support multi-concept queries. We find that adding LoRA representations is effective and outperforms orthogonal adaptation~\cite{po2024orthogonal}. 
Finally, we demonstrate that \methodname achieves state-of-the-art accuracy on the DeepFashion2~\cite{cohen2022my} and ConCon-Chi~\cite{rosasco2024concon} benchmarks, improving prior performance by $4\%-22\%$.

\section{Related Work}
\label{sec:formatting}

\paragraph{Personalized Vision-Language Retrieval.}
Cohen \etal introduced the task of Personalized Vision-Language Retrieval~\cite{cohen2022my} and proposed PALAVRA, a textual inversion approach for the task. PALAVRA first learns a textual inversion network on COCO, which takes a set of images of a concept and predicts an initial pseudo-word token to represent the concept in text queries. This token is then further optimized via backpropagation through CLIP's text encoder. Korbar \etal~\cite{korbar2022personalised} propose a similar approach for the setting of retrieving specific people in videos. \new{A few recent works explore learning embeddings to represent personal concepts for VLM captioning and QA tasks~\cite{alaluf2025myvlm, nguyen2024yo}.}  Yeh \etal ~\cite{yeh2023meta} build on the concept of textual inversion by meta-learning a basis for pseudo-tokens using large scale video data. However, they target a different setting where concepts are learned jointly, using the other concepts in the dataset as negative examples. Recently, Rosasco \etal introduced the Concept-Context Chimera dataset (ConCon-Chi) for personalized retrieval, which provides a more rigorous benchmark for assessing retrieval of personal concepts across diverse contexts than prior datasets. In this work, we depart from prior work by representing new concepts as low rank parameter updates \textit{within} the text encoder instead of pseudo-word tokens. We show that our approach more effectively composes personal and general knowledge, while requiring few parameters per-concept.

\paragraph{Personalized Generation.}
Personalized generation is a more studied related task that generates new images of a personal concept using text-to-image diffusion models. Some approaches use textual inversion to optimize pseudo-word tokens to use within text prompts \cite{gal2022image, dong2022dreamartist}, while others like Dreambooth \cite{ruiz2023dreambooth} and Custom Diffusion \cite{kumari2023multi} find that tuning the weights of the diffusion U-net generates personal concepts with better fidelity. Recent work has focused on selectively tuning certain parameters to promote parameter efficiency and speed \cite{dong2022dreamartist,han2023svdiff, kumari2023multi, yuan2023inserting}, with some leveraging low rank constraints \cite{ruiz2024hyperdreambooth,ham2024personalized, po2024orthogonal, tewel2023key}. \new{Most related to our work is Perfusion~\cite{tewel2023key}, which learns rank-one updates to the diffusion U-net with a key-locking mechanism to constrain updates to the concept's spatial location in the feature map.} While personalization via parameter updates has become commonplace for personalized generation, it has not yet been explored for personalized retrieval, motivating our work. Due to the differences in the nature of the tasks (generative vs.\ discriminative) and models (text-to-image diffusion vs.\ dual encoder), we find that retrieval demands a different strategy for applying internal parameter updates; updating even sparse sets of parameters can result in catastrophic forgetting of the model's general knowledge. Instead of updating parameters throughout the full model, we apply a single rank-one parameter update to the final layer of CLIP's text encoder and directly regularize these parameters to avoid catastrophic forgetting of general knowledge.

\paragraph{Composed Image Retrieval.} Another related task is composed image retrieval \cite{vo2019composing}, which takes an image and textual modification as inputs and performs image retrieval. Approaches have leveraged CLIP for composed image retrieval \cite{baldrati2022conditioned, baldrati2022effective}, with Pic2Word \cite{saito2023pic2word} and SEARLE \cite{baldrati2023zero} learning textual inversion networks to predict a pseudo-word token for the input image. Importantly, composed image retrieval differs from personalized retrieval in that it does not require instance-level recognition of the same concept (\eg, retrieving the exact same person), and instead aims to retrieve images portraying a similar semantic class, layout, or style.

\paragraph{Few-Shot Adaptation of Vision-Language Models.} Our task also relates to work on adapting models like CLIP for few-shot classification. Prior works primarily leverage prompt tuning \cite{lester2021power}, which learns input tokens to represent new classes \cite{tsimpoukelli2021multimodal, zhou2022conditional, zhou2022learning, bulat2023lasp, chen2022plot, derakhshani2022variational, lu2022prompt, yao2023visual, zhu2023prompt, xing2023dual}. A few works consider tuning within the encoder by applying prompts to all layers \cite{khattak2023maple, lee2023read} or augmenting the encoder with adapter modules \cite{zhang2022tip,gao2024clip}. An important distinction is that while CLIP's general knowledge may aid with learning classes from few examples, the general knowledge is not required for these downstream classification tasks. In contrast, personalized retrieval requires not just recognizing the personal concept, but composing it with retained general knowledge.

\section{Personalized Low-Rank Adaptation for Retrieval (POLAR)}

\begin{figure}[t]
    \centering
    \includegraphics[width=\linewidth]{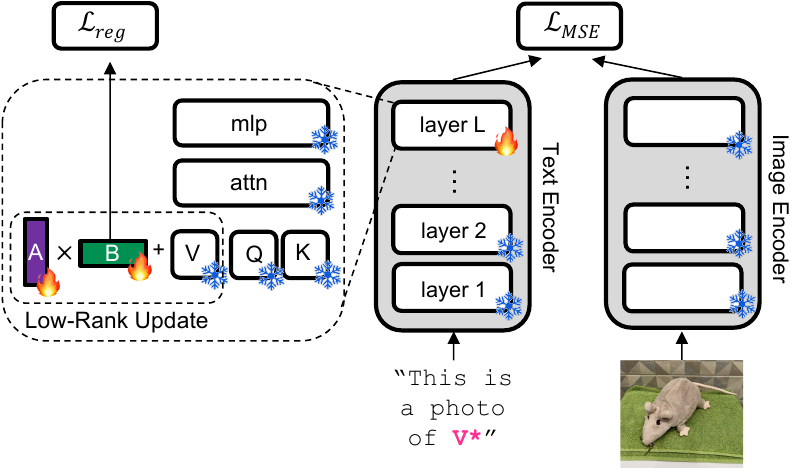}
    \vspace*{-2mm}
    \caption{{\bf Overview of \methodname.} For a personal concept, \methodname learns a rank-1 update to the value transform in the final layer of the text encoder. To maintain general knowledge during personalization, we impose a regularization loss on the update.}
    \vspace*{-2mm}
    \label{fig:method}
\end{figure}

\subsection{Problem Formulation} 
We follow the PerVL problem setup established by Cohen \etal~\cite{cohen2022my}. Given a pretrained vision-language model $\psi$, we aim to learn an adapted model $\psi'$ that is able to recognize a new personal concept $c$ (\eg, ``my coffee mug"). The pretrained model $\psi$ consists of an image encoder $\psi_{I}$ and a text encoder $\psi_{T}$. These encoders map image and textual inputs to a shared embedding space. At personalization time, we are given $N_c$ images $\{I^c_i\}_{i=1}^{N_c}$ depicting concept $c$ as training data. Following Cohen \etal~\cite{cohen2022my}, we are also given the class name $\mathcal{C}_c$ of concept $c$ (\eg, ``mug"). At retrieval time, the input to the model consists of a textual query $q$ and a set of $N_r$ images $\{I^r_i\}_{i=1}^{N_r}$ constituting the retrieval database. We compute the language embedding for an input query $q$ as $\psi'_{T}(q)$, and perform retrieval by computing the cosine similarity between $\psi'_{T}(q)$ and the image embedding for each retrieval image as 
\begin{equation}
    \text{sim}(\psi'_{T}(q), \psi'_{I}(I^r_i))=\frac{\langle\psi'_{T}(q), \psi'_{I}(I^r_i)\rangle}{||\psi'_{T}(q)||_2 || \psi'_{I}(I^r_i)||_2}
    \label{eq:cos_similarity}
\end{equation}
where $\langle\cdot\rangle$ denotes the inner product.
The top retrieval for the query is the image $I^r_i$ corresponding to the embedding with the highest similarity to the query embedding $\psi'_T(q)$.

\subsection{Rank-One Personalized Value Updates}
In contrast to prior work that represents a concept as a learned input token for the text encoder~\cite{cohen2022my, yeh2023meta, baldrati2023zero}, \methodname learns a low-rank \textit{parameter update} to the text encoder for each concept (Fig.~\ref{fig:method}). We leverage LoRA~\cite{hu2021lora}: for a weight $W \in \mathbb{R}^{m\times n}$ in a pretrained model, LoRA learns a low-rank update $\Delta W$, performing a modified forward pass as
\begin{equation}
    y=(W + \Delta W) x = Wx  + BA x
    \label{eq:lora}
\end{equation}
where $B \in \mathbb{R}^{m\times r}$, $A \in \mathbb{R}^{r\times n}$, and $r < \min(m, n)$ is the chosen rank of the weight update. By selecting $r << \min(m, n)$, LoRA tunes minimal parameters in the original model $\psi$. For each concept $c$, we empirically choose to learn a rank-one ($r=1$) update to the value transform of the final attention layer $L$ in $\psi_T$. Our choice of $r=1$ reflects our goal to represent a single concept from very limited examples, while minimally interfering with the model's existing knowledge. Let $Q_L, K_L, V_L \in \mathbb{R}^{d \times d}$ be the pretrained text encoder's query, key, and value transforms for the attention mechanism in the final transformer layer $L$, where $d$ is the internal dimension of $\psi_{T}$. For each attention head, the output of the multi-head attention layer is calculated as

\begin{equation}
    \text{Attention}(x) = \text{softmax}\mathopen{}\left(\frac{Q_L x (K_L x)^T}{\sqrt{d}}\right)\mathclose{}V_{L,c}'x
    \label{eq:attn}
\end{equation}
where $V'_{L,c}$ is the value transform updated with LoRA as:
\begin{equation}
    V'_{L,c} = V_L + B_{L,c}A_{L,c}\,.
    \label{eq:value_lora}
\end{equation}

$V'_{L,c}$ is learned separately for each concept $c$. Following prior work on personalized generation \cite{kumari2023multi, ham2024personalized}, we choose a fixed token (e.g. \textit{``sks"}) in CLIP's vocabulary as a placeholder for the concept's place in the input queries. During training, we randomly select a template textual query (\eg ``An image of sks") to associate with each training image to form a set of text-image pairs
$\{q_i^c, I_i^c\}_{i=1}^{N_c}$. We supervise the learning of $V'_{L,c}$ by using a mean-squared error (MSE) loss to push the normalized text and image embeddings for each training pair close together in the embedding space:
\begin{equation}
    \mathcal{L}_\text{MSE}=\frac{1}{N_c}\sum_{i=1}^{N_c} \left(\frac{\psi'_{T,c}(q_i)}{||\psi'_{T,c}(q_i)||_2} - \frac{\psi_I(I^c_i)}{||\psi_I(I^c_i)||_2}\right)^2
    \label{eq:mse_loss}
\end{equation}
where $\psi'_T,c$ represents the text encoder with concept $c$'s LoRA update applied in the forward pass (Eq.~(\ref{eq:attn})).

\subsection{Regularization}
A key challenge for personalized retrieval is to learn update weights that supply necessary personalized information without overriding the model's general knowledge, which is necessary for retrieving the personal concept in different contexts (\eg, ``my dog Fido catching a frisbee'' and ``my dog Fido sitting on the couch''). We therefore propose a regularization scheme for \methodname. Differently than prior works that construct additional regularization examples to use during training~\cite{cohen2022my, baldrati2023zero}, we instead exploit the structure of our low-rank updates to directly minimize updates to the original representation. 
From Eq.~(\ref{eq:value_lora}), our parameter update alters the representation $V_L x$ by adding the term $B_{L,c}A_{L,c}x$. With our use of rank $r=1$, we can interpret $A_{L,c}x$ as computing dot product similarity between the vectors $A_{L,c}^T$ and $x$, which determines the scale of an added directional update $B_{L,c}$. Our regularization comprises two components: first, we add a penalty on the size of the weights in $B_{L,c}$ to avoid unnecessary deviation from CLIP's existing representation, \ie, when the term $B_{L,c}A_{L,c}x = 0$, the text embedding will be the same as CLIP's original representation ($\psi'_T(q) = \psi_T(q)$). We modify our loss by adding a squared-$L_2$ regularization over the weights $B_{L,c}$:
\begin{equation}
    \mathcal{L} = \mathcal{L}_\text{MSE} + \lambda \mathcal{L}_\text{reg}, \quad \mathcal{L}_\text{reg}= |B_{L,c}|^2
    \label{eq:reg_loss}
\end{equation}
where $\lambda$ is a tunable hyperparameter that determines the relative weight of regularization in the loss. Second, we impose the constraint $||A_{L,c}||_2 = 1$, encouraging $A_{L,c}$ to learn to selectively identify when to apply personal context based on directional similarity with the incoming representation $x$, while the magnitude of the personal update is controlled entirely by the regularized $B_{L,c}$.

\subsection{Merging parameters for multi-concept queries}
For queries that reference multiple personal concepts (\eg, ``my dog Fido is playing with Rex's favorite frisbee"), we propose \textit{merging} the parameter updates for the concepts into one weight update that is applied during encoding to provide personal context for both concepts. Let $V'_{L, c_1}$ and $V'_{L, c_2}$ be the individually learned weight updates for concepts $c_1$ and $c_2$. We construct a combined weight update as:
\begin{equation}
    V'_{L, c_1+c_2} = V'_{L, c_1} + V'_{L, c_2}.
    \label{merged_weight}
\end{equation}
During the forward pass, this setup equates to adding the parameter updates for both personal concepts to the representation. This update is also equivalent to constructing a rank $r=2$ update via matrix concatenation along the rank dimension as:
\begin{equation}
V'_{L, c_1+c_2} = \begin{bmatrix}B_{L,c_1} & B_{L,c_2}\end{bmatrix} \begin{bmatrix}A_{L,c_1} \\ A_{L,c_2} \end{bmatrix}.
\end{equation}
This approach also generalizes to merging greater than two concepts. We explore other merging strategies in Tab.~\ref{tab:merging}.

\section{Experiments}
\label{sec:experiments}

In this section, we evaluate \methodname and compare it to existing works. We also provide ablations to give insight into the design choices within our method. We include further experiment details, analysis of personalization time, and discussion of limitations in the supplemental material.

\subsection{Datasets}
\paragraph{DeepFashion2.} Cohen \etal\cite{cohen2022my} define a personalized retrieval benchmark on the DeepFashion2 dataset \cite{ge2019deepfashion2}, where the 50 personal concepts are different clothing items. The test set includes 221 captions and images for retrieval.
\paragraph{ConCon-Chi.} The recent ConCon-Chi dataset aims to more comprehensively evaluate unique personal concepts in a variety of contexts\new{; we thus use it primarily for our analysis}.
It consists of 20 concepts including household objects and chimeric concepts (combinations of multiple objects). \new{There are 1084 context queries (735 single-concept, 349 multi-concept), and 4008 retrieval images.} 
\new{For direct comparisons to the original baselines~\cite{rosasco2024concon}, we report on the full TEST set, which includes 3 validation concepts. We verify our gains hold on the TEST-UNSEEN split, which excludes these concepts, in the supplemental.}

\subsection{Evaluation Protocol}

\paragraph{Context Queries.} Context queries perform retrieval on a caption referencing the personal concept in a particular context (\eg, ``my dog Fido catching a frisbee in the backyard"). The ground truth consists of the images labeled as matching this prompt. For DeepFashion2, there is one ground truth image per context query, while for ConCon-Chi there are 1-130 ground truth images (average $\approx6$).

\paragraph{Concept-only Queries.} Following Yeh \etal~\cite{yeh2023meta}, we also report retrieval accuracy on ``concept-only" queries to evaluate the model's ability to recognize the personal concept independent of context. For each concept, we use the input query ``An image of V*" and compute retrieval metrics where the ground truth is all retrieval images that contain the concept. For ConCon-Chi, we include only single-concept images \new{(2430 images)} in the retrieval database.

\paragraph{Metrics.} We report retrieval accuracy using standard benchmark metrics for the datasets: mean reciprocal rank (\textbf{mRR}) -- the average inverse rank of the first retrieved ground truth image; recall-at-$k$ (\textbf{r@k}) -- the average success rate within the top $k$ retrievals; and mean average precision (\textbf{mAP}) -- the area under the precision-recall curve, averaged over all queries. We report mAP for settings with multiple ground truth images per query (concept-only queries on both datasets, and context queries on ConCon-Chi).

\paragraph{Implementation Details.} For our main method on ConCon-Chi, we use $\lambda=0.35$. We train for 500 iterations with learning rate $0.001$ and the Adam optimizer. Our model converges within 50 epochs. Because we optimize minimal parameters and backpropagate through only the final layer, personalization is fast, taking under 1 second on a V100 GPU. On DeepFashion2 we append the classname to V* (\eg, ``sks dress") like in PALAVRA~\cite{cohen2022my} and use $\lambda=0.1$.  We use the same template prompts as PALAVRA for training. We provide further details in the supplemental.

\begin{table}
    \centering
    \footnotesize
    \begin{tabular}{lc|cc|cc}
        \toprule
        Method & Arch. & \multicolumn{2}{c}{Context} & \multicolumn{2}{c}{Concept-only} \\
        & & mRR & recall@5 & mRR & mAP \\
        \midrule
        Adapter & ViT-B/32 & $5.9$ & - & -& -\\
        COLLIE~\cite{skantze2022collie} & ViT-B/32 & $7.9$ & - & - & - \\
        Text Only & ViT-B/32 & $17.6$ & - & - & - \\
        AvgIm + Text & ViT-B/32 & $18.8$ & - & - & - \\
        PALAVRA~\cite{cohen2022my} & ViT-B/32 & $28.4$ & $39.2$ & - & - \\
        SEARLE~\cite{baldrati2023zero} & ViT-B/32  & $21.90$ & $27.15$ & 
         $25.97$ & $12.74$ \\
        Ours & ViT-B/32 & $\textbf{34.82}$ & $\textbf{44.88}$ & $\textbf{59.26}$ & $\textbf{28.75}$ \\
        \hline
        SEARLE~\cite{baldrati2023zero} & ViT-L/14 & $27.62$ & $34.12$ & $32.07$ & $16.17$ \\
        Ours & ViT-L/14 & $\textbf{40.72}$ & $\textbf{51.31}$ & $\textbf{65.96}$ & $\textbf{35.07}$ \\
        \bottomrule
    \end{tabular}
    \vspace*{-2mm}
    \caption{Comparison to prior work on the DeepFashion2 retrieval benchmark with 5 training images per concept. We report the mean over over 5 runs with 5 randomly chosen training images of the concept per run (see supplemental for standard error).}
    \label{tab:df2_main}
    \vspace{-1em}
\end{table}

\begin{table}
\begin{subtable}[t]{1.0\linewidth}
    \centering
    \footnotesize
    \begin{tabular}{l|ccc|cc}
        \toprule
        Method & \multicolumn{3}{c}{Context} & \multicolumn{2}{c}{Concept-only} \\
        & mRR & mAP & recall@1 & mRR & mAP \\
        \midrule
        \textit{Coarse (class name)} & 24.21 & 16.83 & 14.48 & - & - \\
        \textit{Discriminative}$^\dagger$ & 43.16 & 30.16 & 31.92 & - & - \\
        \textit{Rich}$^\dagger$ & 40.58 & 27.65 & 29.98 & - & - \\
        \midrule
        PALAVRA \cite{cohen2022my} & 35.99 & 23.59 & 26.75 & - & - \\
        Pic2Word \cite{saito2023pic2word} & 38.62 & 26.39 & 27.68 & - & - \\
        SEARLE \cite{baldrati2023zero} & 43.93 & 30.74 & 33.49 & 96.67 & 61.94 \\
        Ours & \bf 46.33 & \bf 32.33 & \bf 36.16  & \bf 100.00 & \bf 68.71 \\
        \bottomrule
    \end{tabular}
    \caption{Comparison to prior work on the ConCon-Chi benchmark.}
\end{subtable}
\begin{subtable}[c]{1.0\linewidth}
    \centering
    \footnotesize
    \begin{tabular}{l|ccc|ccc}
        \toprule
         & \multicolumn{3}{c}{Context (Single-concept)} & \multicolumn{3}{c}{Context (Multi-concept)} \\
        Method & mRR & mAP & r@1 & mRR & mAP & r@1 \\
        \midrule
        SEARLE & 49.50 & 35.25 & 39.05 & 32.06 & 21.22 & 21.78 \\
        Ours & \bf 51.64 & \bf 36.73 & \bf 41.77 & \bf 35.13 & \bf 23.05 & \bf 24.36 \\
        \bottomrule
    \end{tabular}
    \caption{Results for ConCon-Chi single-concept vs. multi-concept queries.}
\end{subtable}
\label{tab:ccc_main}
\vspace*{-2mm}
\caption{Our approach achieves state of the art results on the challenging ConCon-Chi benchmark on all metrics. We also break down the results of our method and SEARLE\cite{baldrati2023zero} by single-concept and multi-concept queries, demonstrating best results on both. $\dagger$ refer to ConCon-Chi's provided text descriptors for each concept, which serve as oracles since they use knowledge of all concepts to manually determine a differentiating description.}
\vspace*{-2mm}
\label{tab:ccc_main}
\end{table}

\subsection{Comparison to Prior Work}
We compare our method to prior work on DeepFashion2 in Tab.~\ref{tab:df2_main} and ConCon-Chi in Tab.~\ref{tab:ccc_main}. For DeepFashion2 we compare against PALAVRA and the baselines reported in its paper \cite{cohen2022my}. We also run SEARLE~\cite{baldrati2023zero}, a zero-shot composed image retrieval network, using the publicly available checkpoints. Our method achieves state-of-the-art results in the standard setting using the CLIP ViT-B/32 architecture. We also see improvement over SEARLE when using the larger CLIP ViT-L/14 architecture.

\begin{figure*}[t]
  \centering
\includegraphics[width=1.0\linewidth]{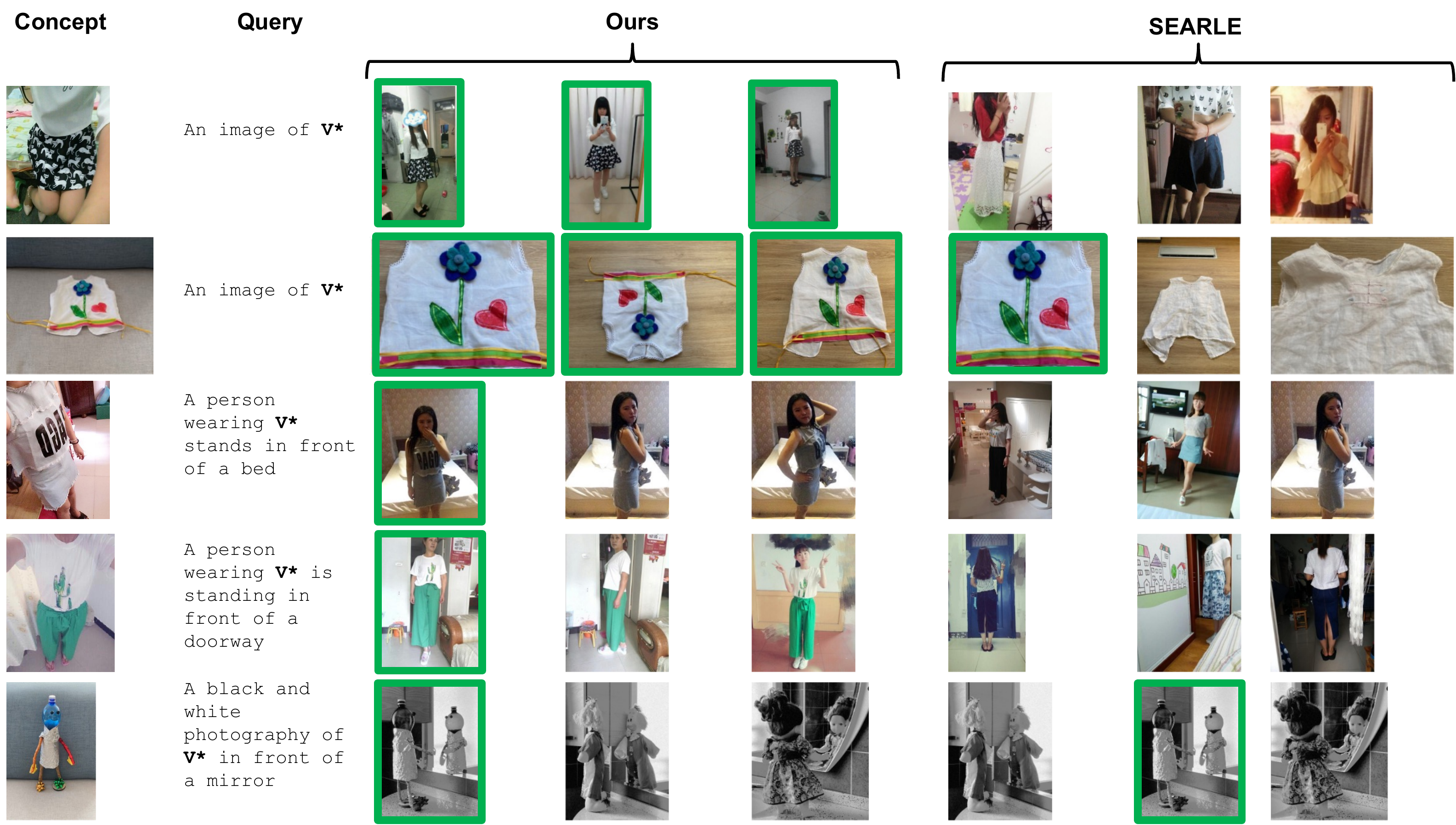}
    \vspace*{-4mm}
   \caption{We compare the top 3 retrievals of our method vs. SEARLE for personal queries in the ConCon-Chi and DeepFashion2 datasets, with green borders indicating correct retrievals. We observe that SEARLE struggles to differentiate between concepts of similar classes, such as different clothing items. Our method more consistently retrieves the correct concept in the correct context, demonstrating effective composition of personal and general knowledge.}
   \label{fig:qualitative}
   \vspace{-1em}
\end{figure*}

On ConCon-Chi, we compare against the baselines reported in the benchmark's paper~\cite{rosasco2024concon}. All methods use the CLIP ViT-L/14 architecture. Differently from DeepFashion2, the zero-shot composed image retrieval methods (SEARLE and Pic2Word) outperform PALAVRA, which suggests that while they are effective in some settings, they struggle with differentiating between several similar concepts like the clothing items in DeepFashion2. This is qualitatively demonstrated in Fig.~\ref{fig:qualitative}, which compares our method's retrievals with SEARLE. Our method, however, also achieves state-of-the-art results on ConCon-Chi, demonstrating flexibility across different benchmarks. Additionally, we achieve stronger mAP for the concept-only queries task than SEARLE.

\paragraph{Evaluating General Knowledge in LoRA vs.\ Token Learning.}
Our primary hypothesis is that learning a small, regularized parameter update within the encoding process can more effectively allow CLIP to reason about personal and general information together. While concept-only queries assess personal knowledge and context queries assess the combination of personal and general knowledge, our setting of tuning the text encoder's parameters also allows us to measure the retention of general knowledge. We do so by inputting \textit{general queries}, which do not reference the personal concept, to the text encoder with our parameter update for the concept still applied. To source non-personal queries, we use LLaVA~\cite{liu2024visual} to caption each image in the retrieval set. We define a new metric \textbf{VLM caption recall@10}, which measures the success rate of retrieving the image for which the caption was generated in the top 10 retrievals. We choose 10 retrievals because ConCon-Chi has many similar images for which the same caption is valid. A drop from the original CLIP's performance on this metric (52.69) indicates forgetting of general knowledge.

\begin{table}
    \centering
    \footnotesize
    \begin{tabular} {l|cc|cc|c}
        \toprule
        Method & \multicolumn{2}{c}{Context \new{(Single)}} & \multicolumn{2}{c}{Concept-only} & VLM cap \\
        & mRR & mAP & mRR & mAP & r@10 \\
        \midrule
        Original CLIP & 29.39 & 20.76 & 10.75 & 6.46 & \bf 52.69 \\
        Text. Inv. (1 tok) & 42.45 & 32.93 & 97.50 & 64.71 & N/A \\
        Text. Inv. (2 tok) & 41.73 & 27.94 & {\bf 100.00} & 63.88 & N/A \\
        Prompt  (1 tok) & 31.77 & 20.70 & 96.25 & 58.95 & 30.84 \\
        Prompt (2 tok) & 33.14 & 20.93 & {\bf 100.00} & 64.49 & 15.35 \\
        \new{Text. Inv. + Ours} & \new{39.29} & \new{26.55} & \new{\bf 100.00} & \new{64.72} & 52.57 \\
        \midrule
        Ours & {\bf 51.64} & {\bf 36.73} & {\bf 100.00} & {\bf 68.71} & {52.62} \\
        \bottomrule
    \end{tabular}
    \vspace*{-2mm}
    \caption{We compare our approach, which updates the weights of the text encoder, to tuning input tokens on ConCon-Chi. Rows 2-3 represent Textual Inversion, where a learned token is applied in place of the personal concept in queries that reference the concept. Rows 4-5 tune prompt tokens that are prepended to all text queries. \new{Row 6 learns both our parameter update and the token}. Tuning tokens can achieve competitive results on concept-only queries but struggles on contextual queries that require composing personal and general knowledge. This catastrophic forgetting of general knowledge is reflected by our VLM caption matching metric. We propose a new approach that achieves strong performance on contextual, concept-only, and general queries.}
    \label{tab:token_tuning}
    \vspace*{-4mm}
\end{table}

\begin{figure*}[t]
  \centering
\includegraphics[width=1.0\linewidth]{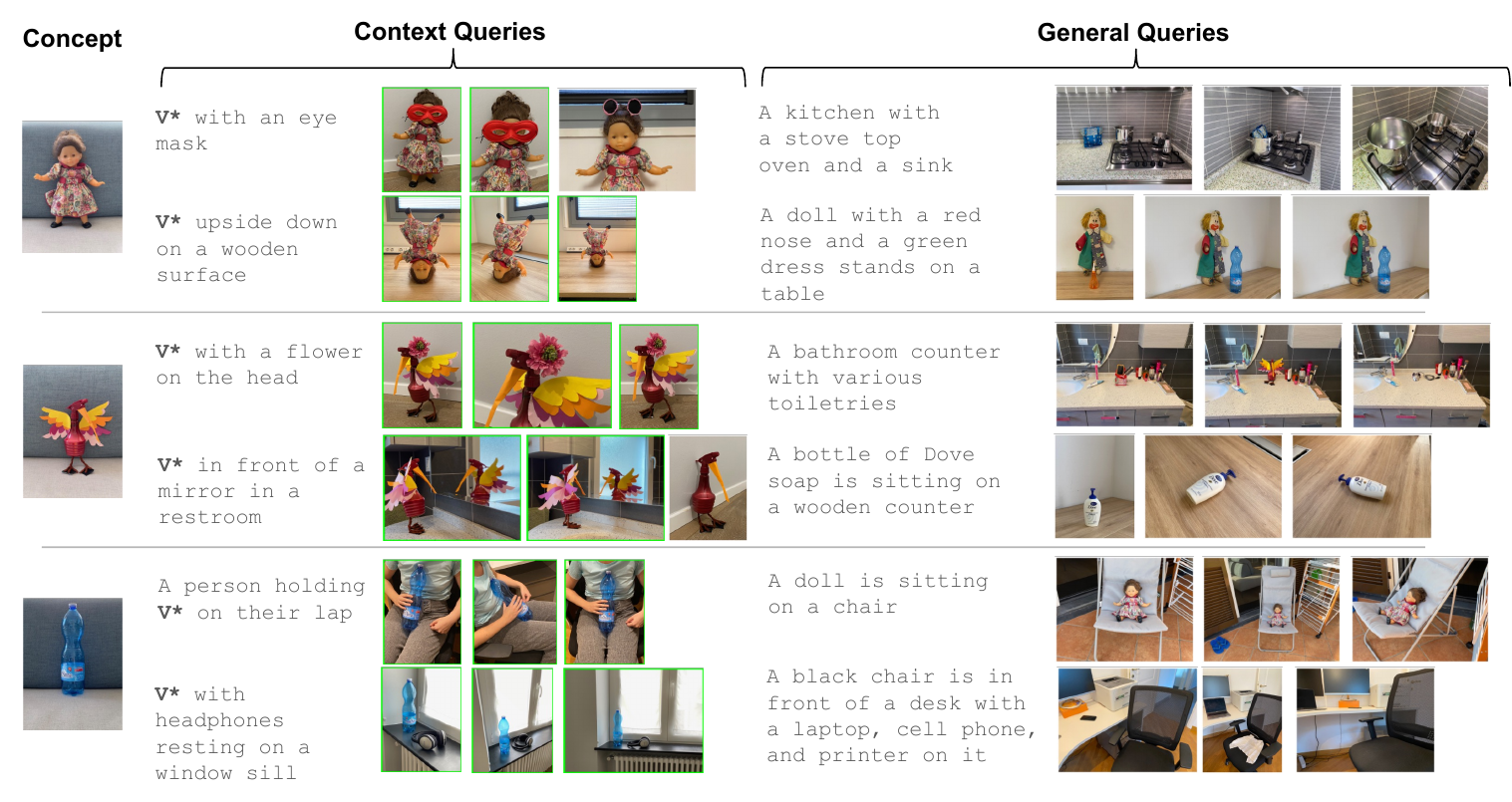}
    \vspace*{-8mm}
   \caption{Our parameter updates enable personalized retrieval without overriding the model's general knowledge. On the left we show the top 3 retrievals of our personalized model on context queries referencing the personal concept, with green indicating ground truth correct retrievals. On the right, we show the results of querying the same personalized model for general VLM-generated captions. There is not exhaustive ground truth for which images match each caption; however, qualitatively our model retrieves appropriate images in all cases.}
   \vspace{-1em}
   \label{fig:context_general}
\end{figure*}

In Tab.~\ref{tab:token_tuning}, we compare our approach of learning internal parameter updates with learning input tokens in the same training setting. We consider Textual Inversion (TI), which learns a token that is integrated into input queries via a pseudo-word for the concept (\eg ``A photo of V* jumping), and Prompt Tuning \cite{zhou2022learning}, \new{which bridges the gap between TI and parameter updates} by prepending learned prompt tokens to all queries (both personal and general). In contrast to TI, this allows us to use our VLM caption metric to measure the interference of the prompt tokens on general queries. We includes results with 1 learned token as well as 2 learned tokens, which is equivalent in size to our rank-1 parameter update. With prompt tuning, we can achieve strong results on concept-only queries, but our VLM caption metric shows that this strength comes at the cost of catastrophic forgetting of general knowledge. \new{We also combine TI with our method by learning the input token in addition to our parameter update. We observe that learning the token reduces context performance, suggesting that the inserted learned token interferes with general knowledge.} In contrast, our parameter updates achieve strong performance on personal queries while not interfering with CLIP's general knowledge, even when applied to non-personal queries. We hypothesize this is due to the minimally invasive structure of our updates, and that they are applied late in the encoding process in contrast to learned input tokens which influence the entire encoding process. We qualitatively illustrate this finding in Fig.~\ref{fig:context_general}; with the parameter update for a personal concept applied, the model successfully performs retrieval for both personalized queries and general queries.

\subsection{Ablation Study}
A key choice in developing our method is determining an effective, minimal set of parameters within the model for which to apply personalized updates. In this section, we empirically investigate rank size of the LoRA updates, which layers in the encoding process to apply LoRA updates, and on which parameters to learn LoRA updates. We additionally ablate our regularization and multi-concept merging strategies. All experiments use the CLIP ViT-L/14 architecture on the ConCon-Chi dataset with 5 training images, and we report our results on single-concept context queries (we ablate the merging strategy for multi-concept queries in Sec.~\ref{sec:merging}). We performed our model selection (design choices for our main method and regularization weight $\lambda$) on the validation split (3 concepts) and report our main results on the test split.

\paragraph{LoRA Rank.} We ablate the rank of the LoRA updates in Tab.~\ref{tab:lora_rank}, which controls the number of learned parameters for each update. We observe only small retrieval accuracy gains on single-concept queries as the rank increases, and achieve better concept-only and competitive multi-concept results with rank=1. The rank-1 concept update is also the most parameter efficient, storing only $2d$ parameters per concept, where $d$ is the encoder's internal dimension.

\begin{table}
    \centering
    \footnotesize
    \begin{tabular}{l|ccc|cc|c}
        \toprule
        LoRA & \multicolumn{3}{c}{Context (Single-Concept)} & \multicolumn{2}{c}{Concept-only} & \multicolumn{1}{c}{VLM cap} \\
        rank & mRR & mAP & r@1 & mRR & mAP & r@10 \\
        \midrule
        r=2 & \bf 52.31 & 36.59 & \bf 42.04 & \bf 100.00 & 66.07 & \bf 52.78  \\
        r=4 & 51.52 & 36.60 & 41.36 & \bf 100.00 & 68.13 & 52.61 \\
        r=8 & 51.49 & 36.58 & 41.36 & \bf 100.00 & 68.15 & 52.62 \\
        r=16 & 51.67 & 36.66 & 41.50 & \bf 100.00 & 67.93 & 52.62 \\
        \midrule
        r=1 & 51.64 & \bf 36.73 & 41.77 & \bf 100.00 & \bf 68.71 & 52.62 \\
        \bottomrule
    \end{tabular}
    \vspace*{-2mm}
    \caption{Ablation of LoRA rank on ConCon-Chi.}
    \label{tab:lora_rank}
\end{table}

\begin{table}
    \footnotesize
    \begin{tabular}{l|ccc|cc|c}
        \toprule
        Layer(s) & \multicolumn{3}{c}{Context (Single-Concept)} & \multicolumn{2}{c}{Concept-only} & VLM cap \\
        & mRR & mAP & r@1 & mRR & mAP & r@10 \\
        \midrule
        11,12 & 50.18 & 35.34 & 39.73 & \bf 100.00 & 64.09 & \bf 52.64  \\
        10,11,12 & 51.51 & 35.81 & 41.63 & \bf 100.00 & 65.87 & 52.62  \\
        all layers & 43.23 & 29.93 & 32.52 & 97.50 & 63.77 & 52.45 \\
        1 & 44.69 & 31.12 & 34.15 & 97.50 & 64.66 & 52.18 \\
        \midrule
        12 & \bf 51.64 & \bf 36.73 & \bf 41.77 & \bf 100.00 & \bf 68.71 & 52.62 \\
        \bottomrule
    \end{tabular}
    \vspace*{-2mm}
    \caption{Ablation of LoRA layers on ConCon-Chi.}
    \vspace*{-2mm}
    \vspace*{-2mm}
    \label{tab:lora_layers}
\end{table}

\paragraph{Architecture Layers.} Tab.~\ref{tab:lora_layers} investigates which transformer layers to apply LoRA updates to, dictating how early or late into the encoder process personalized information is injected. We achieve the strongest results by applying personalization in the final layer (layer 12). We hypothesize that updating later layers is better than earlier layers because it allows our approach to apply a small, targeted update to the developed text query representation to inject personal information. In contrast, earlier layer updates are more likely to alter the full representation - not just the parts semantically belonging to the personal concept. Additionally, our findings align with works that suggest that the later layers in transformer encoders are the most important in constructing the final representation~\cite{gandelsman2023interpreting}. We also do not observe an overall benefit by learning LoRA updates for earlier layers in addition to the final layer; the final layer alone achieves best performance while also requiring the fewest learned parameters per concept. 

\begin{table}
    \footnotesize
    \begin{tabular}{l|ccc|cc|c}
        \toprule
        Param(s) & \multicolumn{3}{c}{Context (Single-Concept)} & \multicolumn{2}{c}{Concept-only} & VLM cap \\
        & mRR & mAP & r@1 & mRR & mAP & r@10 \\
        \midrule
        Q & 16.65 & 11.49 & 7.62 & 32.66 & 10.91 & 51.84 \\
        K & 15.55 & 11.45 & 6.53 & 28.00 & 9.28 & 52.12 \\
        O & 46.99 & 31.04 & 38.50 & 97.50 & 60.98 & 52.52 \\
        Q,K,V,O & 47.52 & 31.69 & 38.78 & 97.50 & 60.90 & 52.54 \\
        Q,V & 51.50 & 35.60 & 41.50 & \bf 100.00 & 65.60 & \bf 52.65 \\
        MLP1 & 43.04 & 27.91 & 33.88 & \bf 100.00 & 55.87 & 52.05 \\
        MLP2 & 49.40 & 32.87 & 38.91 & \bf 100.00 & 55.70 & 51.65 \\
        final proj & 51.15 & 34.98 & 40.82 & \bf 100.00 & 60.58 & 51.58 \\
        \midrule
        V & \bf 51.64 & \bf 36.73 & \bf 41.77 & \bf 100.00 & \bf 68.71 & 52.62 \\
        \bottomrule
    \end{tabular}
    \vspace*{-2mm}
    \caption{Ablation of LoRA parameters on ConCon-Chi.}
    \label{tab:lora_params}
    \vspace*{-2mm}
\end{table}

\paragraph{Parameters.}
Tab.~\ref{tab:lora_params} ablates the component of the transformer layer on which the LoRA is learned. We consider the linear transforms within the attention mechanism (query, key, value, and output), the 2 MLP layers, as well as the model's final projection layer. We see that the query and key transforms alone are not effective for personalization, indicating that directly transforming the output representation is crucial for personalization; re-weighting the existing tokens via altering the attention weight computation is not enough. The strongest results are achieved by updating the value transform. Interestingly, this setting outperforms updating the output transform and the following MLP layers. This result suggests that the value transform's placement in transforming the output of each attention head is optimal for personalization as opposed to later linear transforms that operate after the output of the attention heads is aggregated. We see negligible gains by pairing updates on the value transform with other parameters; in fact, this setting decreases the results on some metrics. Updating the final projection alone also produces competitive results, but lags on the concept-only metrics. Our analysis demonstrates that the value transform within the final layer is optimally positioned to learn a small, targeted update for personalization.

\paragraph{Regularization.}
We ablate our regularization strategy in Tab.~\ref{tab:regularization}. Without regularization, we see drops on the contextual query performance and VLM caption metric, indicating forgetting of general knowledge. The concept-only mAP slightly increases, showing prioritization of personalization over retaining general knowledge. Our regularization strategies are complementary: using both together produces the best results. With our regularization scheme, we are able to produce strong results on contextual queries, avoid any degradation on general caption-matching performance, and still produce good concept-only performance.

\begin{table}
    \centering
    \footnotesize
    \begin{tabular}{p{0.4cm}p{0.4cm}|ccc|cc|c}
        \toprule
        \multicolumn{2}{c}{Reg.} & \multicolumn{3}{c}{Context \new{(Single-Concept)}} & \multicolumn{2}{c}{Concept-only} & VLM cap \\
         A & B & mRR & mAP & r@1 & mRR & mAP  & r@10 \\
        \midrule
        & & 22.51 & 14.35 & 14.83 & \bf 100.00 & \bf 69.89 & 52.52 \\
        \checkmark & & 33.77 & 22.29 & 25.44 & \bf 100.00 & 69.63 & 52.58 \\
        & \checkmark & 39.84 & 26.45 & 31.56 & \bf 100.00 & 69.01 & 52.57 \\
        \checkmark & \checkmark & \bf 51.64 & \bf 36.73 & \bf 41.77 & \bf 100.00 & 68.71 & \bf 52.62  \\
        \bottomrule
    \end{tabular}
    \vspace*{-2mm}
    \caption{Ablation of our regularization scheme: A denotes imposing the constraint $||A_{L,c}||_2 = 1$, and B denotes applying the squared L2 penalty to $B_{L,c}$ (Eq.~\ref{eq:reg_loss}). Without our regularization, the personalized parameter updates cause the model overfit to the concept, producing high concept-only metrics but catastrophically forgetting general knowledge, as reflected in the context and VLM caption metrics. Our regularization prevents this forgetting while still achieving high concept-only performance.}
    \vspace*{-3mm}
    \label{tab:regularization}
\end{table}

\paragraph{Multi-Concept Merging.}
\label{sec:merging}
We consider alternative strategies for merging the low-rank parameter updates of \new{different} concepts in Tab.~\ref{tab:merging}. Our hypothesis is that because we learn a single, constrained update for each concept, the updates for different concepts will be sufficiently different to not interfere with each other. Our results validate this; we see applying each update is better than altering them via averaging or pooling them together. We also consider Orthogonal Adapation~\cite{po2024orthogonal}, which constrains updates for different concepts to be orthogonal to eachother, thus avoiding interference. In this method, the $A$ matrix in the low-rank update is frozen and drawn from a shared orthogonal subspace, while the $B$ matrix is learned. The updates for different concepts are merged by adding them together. Interestingly, we find Orthogonal Adaptation to be less effective than our approach; we hypothesize this is due to the differences between where low rank updates are applied in our method vs. the original Orthogonal Adaptation method, which operates on text to image diffusion models. Whereas in personalized generation, parameter updates are applied throughout the full model, we find learning only a single parameter update is better suited for retrieval. Because we only learn this one parameter update late in the model, certain randomly selected $A$ matrices are not effective for personalization depending on how they interact with the incoming feature representation at that point in the model. Specifically, in computing $B_{L,c}A_{L,c}x$ if the randomly chosen $A_{L,c}$ is orthogonal to $x$, this will eliminate the parameter update altogether. While personalized generation approaches avoid this by adapting the representation throughout all layers of the model, in our setting we find it is better to learn both $A$ and $B$ instead of imposing orthogonality.

\begin{table}
    \centering
    \footnotesize
    \begin{tabular}{l|ccc}
        \toprule
        Merge strategy & mRR & mAP & r@1 \\
        \midrule
        Avg LoRAs & 25.03 & 15.51 & 15.76 \\
        Max LoRAs & 25.27 & 15.56 & 16.05 \\
        Orthogonal Adaptation  \cite{po2024orthogonal} & 28.38 & 19.14 & 18.62 \\
        \midrule
        Add LoRAs & \bf 35.13 & \bf 23.05 &\bf 24.36 \\
        \bottomrule
    \end{tabular}
    \vspace*{-2mm}
    \caption{Performance of different merging strategies for multi-concept queries in ConCon-Chi.}
    \vspace{-3mm}
    \label{tab:merging}
\end{table}

\section{Conclusion}
In this work, we show that updating the internal representation of the CLIP text encoder serves as a better alternative to textual inversion for personalized search. By constraining the parameter update for each concept to a single rank-one update in the value transform of the final layer and strategically regularizing the parameters, we demonstrate that our approach effectively personalizes from a few image examples while maintaining the model's general knowledge.

\clearpage
\paragraph{Acknowledgments} The authors thank the members of the Hoffman Lab at Georgia Tech and Jitesh Jain for their feedback on this work. Fiona Ryan is supported by the NSF Graduate Research Fellowship under Grant No. DGE-2039655. Any opinion, findings, and conclusions or recommendations expressed in this material are those of the authors(s) and do not necessarily reflect the views of the National Science Foundation.

{
    \small
    \bibliographystyle{ieeenat_fullname}
    \bibliography{main}
}
\clearpage
\clearpage
\setcounter{page}{1}
\maketitlesupplementary

\new{
\section{Results on ConCon-Chi TEST-UNSEEN split}
In order to compare to the baselines reported in the original ConCon-Chi paper~\cite{rosasco2024concon}, we report results on the full TEST split, which contains 3 validation concepts and 17 unseen concepts. However unlike zero-shot methods like SEARLE, we use these 3 validation concepts to select the $\lambda$ regularization hyperparameter. We evaluate on the TEST-UNSEEN split in Tab.~\ref{tab:test-unseen}, which excludes these validation concepts. Our results verify that our accuracy gains hold for the concepts for which $\lambda$ was not tuned.
}

\begin{table}[h]
    \label{tab:test-unseen}
    \scriptsize
    \centering
    \new{
    \begin{tabular}{l|ccc|cc}
        \midrule
        Method & \multicolumn{3}{c}{Context} & \multicolumn{2}{c}{Concept-only} \\
        \vspace{-0.2em}
        & mRR & mAP & recall@1 & mRR & mAP \\
        \midrule
        SEARLE & 43.88 & 30.73 & 33.49 & 96.67 & 61.94 \\
        Ours & \bf 46.17 & \bf 31.99 & \bf 36.29  & \bf 100.00 & \bf 70.65 \\
        \midrule
    \end{tabular}
    \caption{Performance on the TEST-UNSEEN split of ConCon-Chi.}
    \label{tab:test-unseen}
    }
\end{table}

\begin{table*}[]
    \centering
    \footnotesize
    \begin{tabular}{lc|cc|cc}
        \toprule
        Method & Arch. & \multicolumn{2}{c}{Context} & \multicolumn{2}{c}{Concept-only} \\
        & & mRR & recall@5 & mRR & mAP \\
        \midrule
        Adapter & ViT-B/32 & $5.9 \pm 0.7$ & - & -& -\\
        COLLIE~\cite{skantze2022collie} & ViT-B/32 & $7.9 \pm 0.7$ & - & - & - \\
        Text Only & ViT-B/32 & $17.6 \pm 0.0$ & - & - & - \\
        AvgIm + Text & ViT-B/32 & $18.8 \pm 0.4$ & - & - & - \\
        PALAVRA~\cite{cohen2022my} & ViT-B/32 & $28.4 \pm 0.7$ & $39.2 \pm 1.3$ & - & - \\
        SEARLE~\cite{baldrati2023zero} & ViT-B/32  & $21.90\pm 0.39$ & $27.15\pm0.57$ & $25.97\pm0.80$ & $12.74\pm0.48$ \\
        Ours & ViT-B/32 & $\bf 34.82 \pm 0.52$ & $\bf 44.88 \pm 1.17$ & $\bf 59.26 \pm 1.64$ & $\bf 28.75 \pm 0.74$ \\
        \hline
        SEARLE~\cite{baldrati2023zero} & ViT-L/14 & $27.62\pm0.26$ & $34.12\pm0.39$ & $32.07\pm0.90$ & $16.17\pm0.62$ \\
        Ours & ViT-L/14 & $\bf 40.72 \pm 0.27$ & $\bf 51.31 \pm 0.78$ & $\bf 65.96 \pm 0.36$ & $\bf 35.07 \pm 0.65$ \\
        \bottomrule
    \end{tabular}
    \caption{Results from Tab. 1 (main text, comparison on the DeepFashion2 test set) with standard error reported over 5 runs.}
    \label{tab:df2_supp}
\end{table*}

\section{Standard Error on DeepFashion2}
We report the mean and standard error over 5 runs with different random seeds on the DeepFashion2 test set in Tab.~\ref{tab:df2_supp} with 5 randomly selected train images for each concept per run.

\begin{table*}[]
    \centering
    \footnotesize
    \begin{tabular}{ll|ccc|cc|c}
        \toprule
        LoRA & Reg. & \multicolumn{3}{c}{Context (Single-Concept)} & \multicolumn{2}{c}{Concept-only} & \multicolumn{1}{c}{VLM cap} \\
        rank & weight & mRR & mAP & r@1 & mRR & mAP & r@10 \\
        \midrule
        r=2 & $\lambda$=2 & 52.71 & 37.30 & 41.43 & \bf 100.00 & 57.21 & \bf 52.61  \\
        r=4 & $\lambda$=6 & 52.51 & 37.20 & \bf 42.45  & \bf 100.00 & \bf 57.54 & 52.50 \\
        r=8 & $\lambda$=24 & 52.52 & 37.20 & \bf 42.45 & \bf 100.00 & \bf 57.54 & 52.51 \\
        r=16 & $\lambda$=100 & 52.62 & 37.34 & 41.45 & \bf 100.00 & 57.53 & 52.48 \\
        \midrule
        r=1 & $\lambda$=0.35 & \bf 52.75 & \bf 37.82 & 41.51 & \bf 100.00 & 57.49 & 52.47  \\
        \bottomrule
    \end{tabular}
    \caption{Validation split performance and regularization weight for ablation of LoRA rank on ConCon-Chi. For each rank, we sweep over different values for $\lambda$ and report the best-performing value.}
    \label{tab:lora_rank_supp}
\end{table*}

\begin{table*}[]
    \centering
    \footnotesize
    \begin{tabular}{ll|ccc|cc|c}
        \toprule
        Layer(s) & Reg. & \multicolumn{3}{c}{Context (Single-Concept)} & \multicolumn{2}{c}{Concept-only} & VLM cap \\
        & weight & mRR & mAP & r@1 & mRR & mAP & r@10 \\
        \midrule
        11,12 & $\lambda$=2 & 52.42 & 37.36 & \bf 42.40 & \bf 100.00 & 56.99 & \bf 52.66  \\
        10,11,12 & $\lambda$=4 & 52.03 & 37.32 & 41.45 & \bf 100.00 & 57.46 & 52.56  \\
        all layers & $\lambda$=40 & 44.45 & 32.46 & 34.91 & 83.33 & 53.23 & 52.37 \\
        1 & $\lambda$=1 & 43.39 & 32.68 & 33.96 & 83.33 & 54.19 & 52.21 \\
        \midrule
        12 & $\lambda$=0.35 & \bf 52.75 & \bf 37.82 & 41.51 & \bf 100.00 & \bf 57.49 & 52.47 \\
        \bottomrule
    \end{tabular}
    \caption{Validation split performance and regularization weight for ablation of LoRA layers on ConCon-Chi. For each layer set, we sweep over different values for $\lambda$ and report the best-performing value.}
    \label{tab:lora_layers_supp}
\end{table*}

\begin{table*}
    \centering
    \footnotesize
    \begin{tabular}{ll|ccc|cc|c}
        \toprule
        Param(s) & Reg. & \multicolumn{3}{c}{Context (Single-Concept)} & \multicolumn{2}{c}{Concept-only} & VLM cap \\
        & weight & mRR & mAP & r@1 & mRR & mAP & r@10 \\
        \midrule
        Q & $\lambda$=0 & 23.17 & 15.09 & 13.21 & 38.89 & 8.77 & 52.15 \\
        K & $\lambda$=0 & 19.82 & 14.93 & 9.43 & 2.36 & 5.81 & 52.11 \\
        O & $\lambda$=100 & 51.22 & 33.69 & \bf 42.45 & 83.33 & 51.69 & 52.62 \\
        Q,K,V,O & $\lambda$=500 & 51.14 & 33.86 & \bf 42.45 & 83.33 & 51.90 & \bf 52.66 \\
        Q,V & $\lambda$=2 & \bf 53.04 & 37.76 & 42.40 & \bf 100.0 & 56.66 & 52.63 \\
        MLP1 & $\lambda$=50 & 44.01 & 28.45 & 33.96 & \bf 100.0 & 48.05 & 51.64 \\
        MLP2 & $\lambda$=200 & 50.57 & 33.12 & 38.68 & \bf 100.0 & 49.81 & 51.25 \\
        final proj & $\lambda$=700 & 52.42 & 35.77 & 39.62 & \bf 100.0 & 53.91 & 51.09 \\
        \midrule
        V & $\lambda$=0.35 & 52.75 & \bf 37.82 & 41.51 & \bf 100.00 & \bf 57.49 & 52.47\\
        \bottomrule
    \end{tabular}
    \caption{Validation split performance and regularization weight for ablation of LoRA parameters on ConCon-Chi. For each parameter set, we sweep over different values for $\lambda$ and report the best-performing value.}
    \label{tab:lora_params_supp}
\end{table*}

\section{Ablation Validation Split Results \& Hyperparameters}

We provide the ConCon-Chi validation split results and the value for the regularization weight hyperparameter $\lambda$ for the ablations reported in the main paper: LoRA rank (Tab.~\ref{tab:lora_rank_supp}, LoRA layers (Tab.~\ref{tab:lora_layers_supp}), and LoRA parameters (Tab.~\ref{tab:lora_params_supp}). We performed our search for the value of $\lambda$ resulting in convergence to the highest accuracy for each setting on the validation split. We selected our final model setting (rank=1, layers=12, parameters=V, $\lambda=0.35$) based on the results of these ablations on the validation split.

\section{Comparison to Yeh \etal~\cite{yeh2023meta}}
Yeh \etal~\cite{yeh2023meta} propose a textual inversion approach for PerVL that meta-learns a per-class basis on large scale data, over which the $V^*$ tokens for new concepts are learned as a linear combination. Both the $V^*$ token and basis are updated at personalization time. Differently from the original PerVL setting \cite{cohen2022my}, the tokens for all concepts in the dataset are learned \textit{jointly}, with the vision-text contrastive loss using images of the other concepts as hard negatives and an additional text-text contrastive loss pushing apart the text embeddings for different concepts. We exclude their method from our main comparisons since this is a different setting than that followed by prior methods. Using the other concepts as hard negatives gives the method an advantage at retrieval time since the retrieval database is composed of images of the concepts in the dataset. For DeepFashion2 in particular, where the concepts are all clothing items and many are visually similar, using the other concepts as negatives helps the model distinguish its representation of each concept from visually similar concepts that will appear in the retrieval database. 

To adapt our method to this setting where hard negatives are provided, we create an additional objective that pushes personal textual queries for the concept being learned away from the image embeddings of other concepts in CLIP space. Specifically we define a \textit{negative loss}, $\mathcal{L}_{neg}$, as a negative MSE loss:
\begin{equation}
    \mathcal{L}_\text{neg}=-\frac{1}{N_c}\sum_{i=1}^{N_c} \left(\frac{\psi'_{T,c}(q_i)}{||\psi'_{T,c}(q_i)||_2} - \frac{\psi_I(I^n_i)}{||\psi_I(I^n_i)||_2}\right)^2
\end{equation}
where for each iteration, $\{I^n_i\}$ consists of $N_c$ sampled training images containing a concept that is \textbf{not} concept $c$. We alter Eq. 6 (main text) to be:
\begin{equation}
    \mathcal{L} = \mathcal{L}_\text{MSE} + \mathcal{L}_\text{neg} + \lambda \mathcal{L}_\text{reg}
    \label{eq:neg_loss}
\end{equation}
Note that this training objective differs from Yeh \etal, which uses a set of contrastive losses between the concepts during joint training. We introduce $\mathcal{L}_\text{neg}$ as a means of accomodating hard negatives with minimal changes to our existing training objective and setting.

\begin{table*}
    \centering
    \footnotesize
    \begin{tabular}{l|cc|cc}
        \toprule
        Method & \multicolumn{2}{c}{Context} & \multicolumn{2}{c}{Concept-only} \\
        & mRR & recall@5 & mRR & mAP \\
        \midrule
        Yeh \etal & $34.4\pm0.7$ & $45.2\pm1.1$  & $69.3\pm1.8$ & $40.0\pm1.0$ \\
        Ours & $34.82 \pm 0.52$ & $44.88 \pm 1.17$ & $59.26 \pm 1.64$ & $28.75 \pm 0.74$ \\
        Ours + negs & $\bf42.23\pm0.23$ & $\bf52.57\pm0.35$ & $\bf69.66\pm0.98$ & $\bf40.65\pm0.59$\\
        \bottomrule
    \end{tabular}
    \caption{Comparison to Yeh \etal~\cite{yeh2023meta}, which uses the other concepts as hard negatives during training. We include our method in the original setting (Ours), and our method adapted to also use negatives (Ours + negs). All results use the ViT-B/32 architecture and report mean and standard error over 5 runs.}
    \vspace{-2em}
    \label{tab:df2_negatives}
\end{table*}

\paragraph{Quantitative Comparison} We provide a quantitative comparison on DeepFashion2 in this setting in Tab.~\ref{tab:df2_negatives}. We use the ViT-B/32 backbone for these experiments and set $\lambda_\text{neg}=1$ and $\lambda_\text{reg}=0.1$. Without having the other concepts as hard negatives, our method naturally has lower concept-only performance, as it does not have the advantage of hard negatives to disambiguate between similar concepts. With the addition of negatives, we achieve similar concept-only performance to Yeh ~\etal, and much higher context performance. These results demonstrate that our method better balances personal knowledge and generic knowledge than Yeh ~\etal's textual inversion based method.

\begin{table*}[t]
    \centering
    \footnotesize
    \begin{tabular}{l|l|ccc|cc}
        \toprule
        \# Train Imgs & Method & \multicolumn{3}{c}{Context} & \multicolumn{2}{c}{Concept-only} \\
        & & mRR & mAP & recall@1 & mRR & mAP \\
        \midrule
        0 & \textit{Coarse (class name)} & 24.21 & 16.83 & 14.48 & - & - \\
        & \textit{Discriminative}$^\dagger$ & 43.16 & 30.16 & 31.92 & - & - \\
        & \textit{Rich}$^\dagger$ & 40.58 & 27.65 & 29.98 & - & - \\
        \midrule
        1 & PALAVRA & $34.39\pm1.68$ & $22.56\pm1.29$ & $24.59\pm1.94$ & - & - \\
        & Pic2Word & $37.15\pm1.76$ & $25.23\pm1.20$ & $26.35\pm1.85$ & - & - \\
        & SEARLE & $41.07\pm0.92$ & $28.16\pm0.55$ & $31.16\pm0.94$ & - & - \\
        & Ours & $\bf44.68\pm0.61$ & $\bf30.99\pm0.48$ & $\bf 34.45\pm0.55$ & $\bf 98.83\pm1.62$ & $\bf 65.10\pm0.96$\\
        \midrule
        5 & PALAVRA \cite{cohen2022my} & 35.99 & 23.59 & 26.75 & - & - \\
        & Pic2Word \cite{saito2023pic2word} & 38.62 & 26.39 & 27.68 & - & - \\
        & SEARLE \cite{baldrati2023zero} & 43.93 & 30.74 & 33.49 & \bf 100.00 & 61.68 \\
        & Ours & \bf 46.33 & \bf 32.33 & \bf 36.16  & \bf 100.00 & \bf 68.71 \\
        \bottomrule
    \end{tabular}
    \caption{Comparison to prior work on the ConCon-Chi benchmark, including the single training image setting. For single image training, we report the mean and standard deviation. Our approach achieves state-of-the-art results in both the 1-image and 5-image settings. $\dagger$ indicates oracle descriptions.}
    \label{tab:single_image}
\end{table*}

\section{Single Training Image Experiments on ConCon-Chi} The original ConCon-Chi paper \cite{rosasco2024concon} also reports results where only  a single training image is used per concept. We report results for our method in this setting in Tab.~\ref{tab:single_image}. We use the same hyperparameters as our main ConCon-Chi experiments where all 5 training images per concept are used. We report the mean and standard deviation over each of the 5 training images. Our method performs best in the single-image setting, and our single-image method even outperforms the other methods when they use all 5 training images. This result demonstrates the effectiveness of \methodname even with a single training image per concept.

\begin{table}[h]
    \footnotesize
    \centering
    \begin{tabular}{lcc}
        \toprule
        Method & Iters & Personalization time (ms) \\
        \midrule
        Text. Inv. (1 tok) & 50 & 1597.62 \\
        Ours & 50 & 219.54 \\
        \midrule 
        Text. Inv. (1 tok) & 500 & 15961.97 \\
        Ours & 500 & 1940.34 \\ 
        \bottomrule
    \end{tabular}
    \caption{Total personalization time for a concept in milliseconds of our method vs. textual inversion.}
    \label{tab:timing}
\end{table}

\section{Personalization Time Analysis}

\methodname is fast to personalize and does not require pretraining. For all experiments in Section 4 (main text), we optimize for 500 iterations to ensure all variants converge; however for our main method setting (rank=1, layers=12, params=V, $\lambda$=0.35), our model converges within 50 iterations. We provide runtime analysis in Tab.~\ref{tab:timing}, showing the full personalization time of our ViT-L/14-based method with 5 training images for a concept on a single NVIDIA V100 GPU. We report the personalization time for both 50 iterations and 500 iterations. Because we backpropagate only through the final layer of the text encoder, our method is significantly faster to optimize than traditional textual inversion.

\section{Additional Implementation Details}
\paragraph{DeepFashion2.} We train our ViT-B/32 model for 50 iterations, and our ViT-L/14 model for 200 iterations. We use the Adam optimizer with learning rate 0.001. We use the token ``sks" as $V^*$.

\paragraph{ConCon-Chi.} We train our ViT-L/14 model for 500 iterations. We use the Adam optimizer with learning rate 0.001. We do not append the classname to $V^*$, because the classnames are less likely to be aligned with the concept. For example, several concepts have the classname ``puppet" as they are animal-like objects created from household materials, but this is unlikely to align with CLIP's concept of ``puppet" based on its pretraining. We use the token ``sks" as $V^*$.

\new{
\section{Evaluation of General Knowledge}
\paragraph{VLM Captions.} To generate the captions for calculating our VLM caption recall@10 metric, we prompt LLaVA-1.5-7B~\cite{liu2024visual} with the image and the prompt ``Caption this image in 1-2 sentences." To assess noise in the captions, we manually checked 100 of the captions, finding 88 accurate, 10 with minor errors, and 2 wrong. The metric is intended to assess the performance delta from original CLIP, so a noisy caption equally affects both methods. We choose a permissive threshold of r@10 because the ground truth is determined as the single image from which the caption is generated, but ConCon-Chi has multiple similar images. Our method performs similarly to CLIP across different thresholds, as shown in Tab.~\ref{tab:threshold}.}

\begin{table}[h!]
    \centering
    \footnotesize
    \new{
    \begin{tabular}{l|cccc}
        \toprule
        Method & r@1 & r@5 & r@10 & r@50 \\
        \midrule
        Original CLIP &13.27 & 39.62 & 52.69 & 78.07 \\
        Ours & 13.39 & 39.61 & 52.62 & 78.07 \\
        \bottomrule
    \end{tabular}
    }
    \caption{Evaluation with different recall thresholds for our VLM caption metric.}
    \label{tab:threshold}
\end{table}

\paragraph{\new{Evaluation on general retrieval task.}}
\new{We also evaluate retention of general knowledge by performing general image retrieval on Flick30k~\cite{young2014image} with the parameter update for a concept applied. We report results in Tab.~\ref{tab:flickr}, showing parity with original CLIP.}

\begin{table}[h!]
    \centering
    \footnotesize
    \new{
    \begin{tabular}{l|ccc}
        \toprule
        Method & r@1 & r@5 & r@10 \\
        \midrule
         Original CLIP & 67.76 & 89.78 & 94.26 \\
         Ours & 68.16 & 89.79 & 94.43 \\
         \bottomrule
    \end{tabular}
    \caption{Evaluation on the Flick30k general image retrieval task.}
    \label{tab:flickr}
    }
\end{table}

\paragraph{\new{Evaluation with ConCon-Chi discriminative captions.}}
\new{The ConCon-Chi dataset also includes \textit{discriminative} descriptions for each concept, which are human-annotated text descriptions that differentiate the concepts from one another (\eg, ``bird sprayer puppet"). These descriptions provide an oracle baseline for the benchmark. We also evaluate retention of general knowledge by evaluating image retrieval on ConCon-Chi where each personal concept's place in the image caption annotations is replaced by the concept's discriminative description. Results are provided in Tab.~\ref{tab:discriminative}, showing similar performance to original CLIP.}

\begin{table}[h!]
    \centering
    \footnotesize
    \new{
    \begin{tabular}{l|ccc}
        \toprule
        Method & r@1 & r@5 & r@10 \\
        \midrule
         Original CLIP & 31.92 & 55.17 & 66.51 \\
         Ours & 31.62 & 54.76 & 66.00 \\
         \bottomrule
    \end{tabular}
    \caption{Evaluation on ConCon-Chi general image retrieval using discriminative concept descriptions in captions.}
    \label{tab:discriminative}
    }
\end{table}

\new{\section{Comparison to Weight Decay}}
\noindent\new{We regularize our personalized parameter updates via the $||A_{L,c}||_2=1$ constraint and imposing a squared-$L_2$ penalty on $B_{L,c}$. This strategy is similar to weight decay, which also encourages learning small weights, but differs in two key aspects. First, weight decay is typically applied to all parameters, while we only impose a penalty on the size of $B_{L,c}$. Second, weight decay is implemented differently, directly subtracting a portion of the weights during the optimizer update. Tab.~\ref{tab:weight_decay} compares our regularization scheme to simply using weight decay with the Adam optimizer (with a tuned value of 1e-4) and the AdamW optimizer with default hyperparameters. These results show that simply using Adam/AdamW struggles both with learning the concept (due to applying weight decay to $A_{L,c}$) and retaining general knowledge.
}

\begin{table}[h!]
    \centering
    \footnotesize
    \new{
    \begin{tabular} {@{}l|ccc|cc|@{}c@{}}
        \midrule
        Method & \multicolumn{3}{c}{Context (Single-Concept)} & \multicolumn{2}{c}{Concept-only} & VLM cap \\
        \vspace{-0.3em}
        & mRR & mAP & r@1 & mRR & mAP & r@10 \\
        \midrule
        Adam + wd & 47.58 & 32.34 & 38.64 & 100.00 & 65.59 & 51.20 \\
        AdamW + wd & 49.61 & 34.08 & 39.46 & 97.50 & 59.72 & 51.24 \\
        Ours & {\bf 51.64} & {\bf 36.73} & {\bf 41.77}  & {\bf 100.00} & {\bf 68.71} & {\bf 52.62} \\
        \midrule
    \end{tabular}
    \caption{Comparison of our regularization strategy with optimizer weight decay.}
    \label{tab:weight_decay}
    }
\end{table}

\section{Generalization of Ablations to DeepFashion2}
While we report our main ablations on the ConCon-Chi dataset, we observe similar trends on DeepFashion2. Tab.~\ref{tab:df2_ablation} shows ablating the parameters on which the LoRA is learned on DeepFashion2 for a single run of 5 training images. We see similar results to ConCon-Chi (Tab.~\ref{tab:lora_params}), with the value transform performing best.

\begin{table}[h!]
    \centering
    \footnotesize
    \begin{tabular}{l|cc|cc}
        \midrule
        Params & \multicolumn{2}{c}{Context} & \multicolumn{2}{c}{Concept-only} \\
        \vspace{-0.3em}
        & mRR & r@5 & mRR & mAP \\
        \midrule
        K & 23.97 & 29.41 & 13.51 & 00.08 \\
        O & 35.15 & 44.80 & 58.51 & 30.55 \\
        Q,V & 36.36 & 47.96 & 60.21 & 32.60 \\
        Q,K,V,O & 35.37 & 45.34 & 60.09 & 32.12 \\
        V & \bf 41.35 & \bf 49.32 & \bf 65.48  & \bf 35.02 \\
        \midrule
    \end{tabular}
    \caption{Ablation of LoRA parameters on DeepFashion2.}
    \label{tab:df2_ablation}
\end{table}

\begin{figure*}
    \centering
    \includegraphics[width=1.0\linewidth]{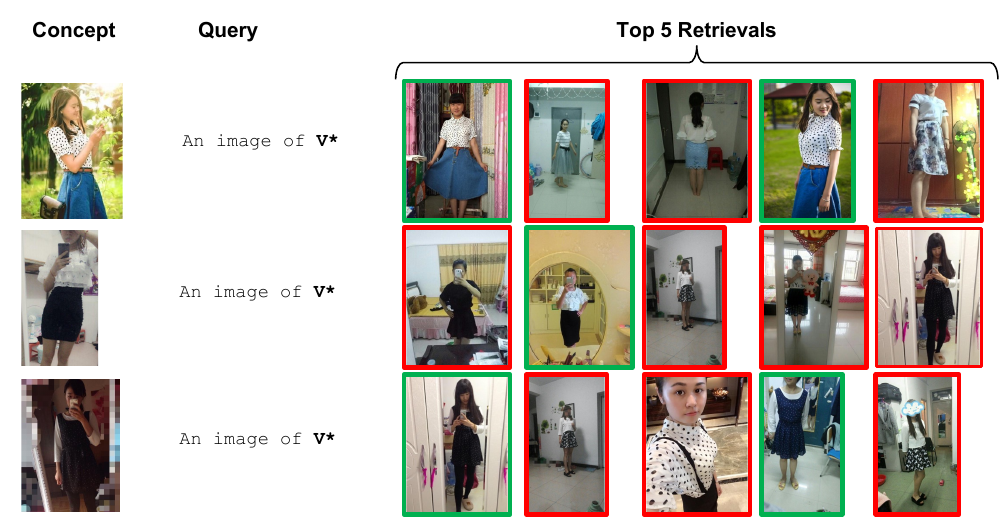}
    \caption{Our method sometimes struggles to differentiate between concepts of the same class with similar visual attributes such as color and pattern. We show concept-only queries from DeepFashion2 where such failures occur, with correct retrievals shown in green and incorrect retrievals shown in red. In row 1, the model retrieves other outfits that also have a white shirt and blue skirt, but the pattern of the shirt differs from the correct concept (\eg, polka dot \vs striped). In row 2, the model fails to disambiguate between black skirts of different shapes. In row 3 where the concept has a black and white polka-dot pattern, the model retrieves some incorrect concepts that also have a black and white polka-dot pattern.}
    \label{fig:df2_failures}
\end{figure*}

\section{Limitations} Like existing approaches in the space of personalized generation that use a fixed $V^*$ token in place of new concepts, we experience sensitivity to the choice of $V^*$. Similar to prior work \cite{kumari2023multi, ham2024personalized, materzynska2023customizing} we find unique single tokens to be the most effective, and we use the token for ``sks" in our main experiments. We observe that selecting a $V^*$ for which CLIP likely has a strong existing representation (\eg, ``dog") makes it more challenging to successfully teach the model the new personalized meaning with limited parameter updates. Future work may explore dynamically determining hyperparameters such as the rank of the LoRA update and the regularization weight for different choices of $V^*$ to eliminate this sensitivity and allow referral to concepts in natural language without the substitution of $V^*$.

Additionally, by updating only the text encoder $\psi_\text{T}$ and not the image encoder $\psi_\text{I}$, our performance is inherently bounded by the frozen image encoder's ability to capture distinguishing visual details. While this choice makes sense practically for our task setting (the image features for all images in the retrieval database can be precomputed by regular CLIP and then the incoming textual queries are encoded by $\psi'_\text{T}$), our approach may struggle to differentiate between visually similar concepts such as different people or objects of the same class. Some works on related tasks avoid this issue by using domain-specific specialized models such as facial feature detectors for personal concepts \cite{korbar2022personalised, alaluf2025myvlm}. However our focus is on minimally adapting CLIP without introducing additional domain-specific models. We show cases where our model fails to distinguish between visually-similar concepts in Fig~\ref{fig:df2_failures}.


\end{document}